%% file: icra_Chablat_Wenger_bis.tex
\documentclass[letterpaper,10pt,twocolumn]{article}
\usepackage[dvips]{graphicx}
\usepackage{subeqn}
\input{macro.tex}

\makeatletter
\def\@normalsize{\@setsize\normalsize{12pt}\xpt\@xpt
\abovedisplayskip 10pt plus2pt minus5pt\belowdisplayskip
\abovedisplayskip
\abovedisplayshortskip \z@ plus3pt\belowdisplayshortskip 6pt plus3pt minus3pt
\let\@listi\@listI}
\def\subsize{\@setsize\subsize{12pt}\xipt\@xipt}
\def\section{\@startsection {section}{1}{\z@}{24pt plus 2pt minus 2pt}
{12pt plus 2pt minus 2pt}{\large\bf}}
\def\subsection{\@startsection {subsection}{2}{\z@}{12pt plus 2pt minus 2pt}
{12pt plus 2pt minus 2pt}{\subsize\bf}}
\makeatother

\font\tencmr=cmr10
\begin{document}
\date{}
\title {\Large\bf The Isoconditioning Loci of A Class of Closed-Chain
Manipulators}
\author{\begin{tabular}[t]{c}
        Damien Chablat \hfill  Philippe Wenger \hfill  Jorge Angeles$^\1$\\
        {\tencmr Institut de Recherche en Cybern\'etique de Nantes} \\
        {\tencmr \'Ecole Centrale de Nantes} \\
        {\tencmr 1, rue de la No\"e, 44321 Nantes, France} \\
        {\tencmr Damien.Chablat@lan.ec-nantes.fr} \hfill
        {\tencmr Philippe.Wenger@lan.ec-nantes.fr} \\
        {\tencmr $^\1$McGill Centre for Intelligent Machines and Department
        of Mechanical Engineering} \\
        {\tencmr McGill University, 817 Sherbrooke Street West} \\
        {\tencmr Montreal, Quebec, Canada H3A 2K6} \\
        {\tencmr Angeles@cim.mcgill.ca}
        \end{tabular}}
\maketitle
\thispagestyle{empty}
\subsection*{\centering Abstract}
{\em The subject of this paper is a special class of closed-chain manipulators. First, we analyze a family of two-degree-of-freedom (dof) five-bar planar linkages. Two Jacobian matrices appear in the kinematic relations between the joint-rate and the Cartesian-velocity vectors, which are called the ``inverse kinematics" and the ``direct kinematics" matrices. It is shown that the loci of points of the workspace where the condition number of the direct-kinematics matrix remains constant, i.e., the isoconditioning loci, are the coupler points of the four-bar linkage obtained upon locking the middle joint of the linkage. Furthermore, if the line of centers of the two actuated revolutes is used as the axis of a third actuated revolute, then a three-dof hybrid manipulator is obtained. The isoconditioning loci of this manipulator are surfaces of revolution generated by the isoconditioning curves of the two-dof manipulator, whose axis of symmetry is that of the third actuated revolute. }
\begin{keyword}
Kinematics, Closed-Loop Manipulator, Hybrid manipulator, Isoconditioning surfaces, Singularity, Working Modes.
\end{keyword}
\section{Introduction}
The aim of this paper is to study (a) a family of two-dof, five-bar
planar linkages and (b) a derivative of this family, obtained when
a third revolute is added in series to the above linkages, with the
purpose of obtaining a three-dof manipulator. For the mechanical
design of this class of manipulators, various features must be
considered, e.g., the workspace volume, manipulability, and
stiffness. The analysis of single-dof closed-loop chains is
classical within the theory of machines and mechanisms
\cite{Hunt:78}. The study of the workspace and the mobility of
closed-loop manipulators, in turn, is given by Bajpai and Roth
\cite{Bajpai:86}. Gosselin \cite{Gosselin:90b}, \cite{Gosselin:90a}
conducted similar analyses for closed-loop manipulators with one
single inverse kinematic solution on both a planar and a spatial
mechanism. One important property of parallel manipulators is that
they admit several solutions to both their inverse and their direct
kinematics. This property leads to two types of singularities.
\par
The singularities of these manipulators are correspondingly
associated with two Jacobian matrices called here the ``inverse
kinematics'' and the ``direct kinematics'' matrices. By means of
the inverse kinematics matrix, we can define the ``working mode''
of the manipulator to separate the inverse kinematics solutions. It
is useful to represent the manipulator in the workspace and to
define its aspects in this workspace. The aspects of a manipulator
are defined in \cite{Chablat:97}. Moreover, a novel three-dof
hybrid manipulator is proposed, which is comparable to the one
proposed by Bajpai and Roth \cite{Bajpai:86}; ours is obtained as
the series array of a one-revolute chain and the two-dof
closed-chain manipulator described above. In this array, the axis
of the former intersects the axes of the two actuated joints of the
latter at right angles.
\par
The  proper operation of a manipulator depends first of foremost on
its design; besides design, the operation depends on suitable
trajectory-planning and control algorithms. In any event, a
performance index needs be defined, whose minimization or
maximization leads to an optimum operation. While various items
come into play when assessing the operation of a manipulator, we
focus here on issues pertaining to manipulability or dexterity. In
this regard, we understand these terms in the sense of measures of
distance to singularity, which brings us to the concept of
condition number \cite{Golub:89}. Here, we adopt the condition
number of the underlying Jacobian matrices as a means to quantify
distances to singularity. Furthermore, we derive the loci of points
of the joint and Cartesian workspaces whereby the condition number
of each of the Jacobian matrices remains constant. For the planar
two-dof manipulators studied here, we term these loci the {\em
isoconditioning curves}, while, for three-dof spatial manipulators,
these curves become the {\em isoconditioning surfaces.}
\section{A Two-DOF Closed-Chain Manipulator}
\begin{figure}[hbt]
  \begin{center}
        \includegraphics[width= 58mm,height= 40mm]{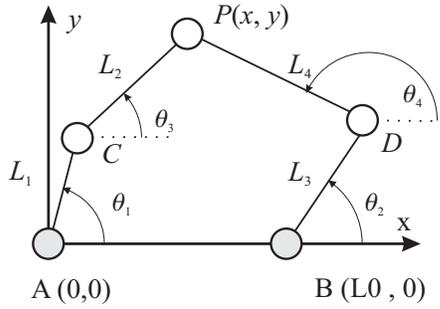}
        \caption{A two-dof closed-chain manipulator}
        \protect\label{figure:manipulateur_general}
  \end{center}
\end{figure}

The manipulator under study  is a five-bar, revolute ($R$)-coupled
linkage, as displayed in Fig.~\ref{figure:manipulateur_general}.
The actuated joint variables are $\theta_1$ and $\theta_2$, while the
Cartesian variables are the ($x$, $y$) coordinates of the revolute
center $P$.
\par
Lenghts $L_0$, $L_1$, $L_2$, $L_3$, and $L_4$ define the geometry of
this manipulator entirely.
However, in this paper we focus on a symmetric manipulators, with $L_1
= L_3$ and $L_2 = L_4$.
The symmetric architecture of the manipulator at hand is justified for
general tasks. In manipulator design, then, one is interested in
obtaining values of $L_0$, $L_1$, and $L_2$ that optimize a given
objective function under some prescribed constraints.
\subsection{Kinematic Relations}
The velocity $\dot{\bf p}$ of point $P$, of position vector \negr p,
can be obtained in two different forms, depending on the direction in
which the loop is traversed, namely,
\begin{subequations}
\begin{equation} 
        \dot{\bf p}= \dot{\bf c} + \dot{\theta}_3 {\bf E}
                     ({\bf p}  - {\bf c} )
        \protect\label{equation:kinematic_1}
\end{equation}
\begin{equation}
        \dot{\bf p}= \dot{\bf d} + \dot{\theta}_4 {\bf E}
                     ({\bf p}  - {\bf d})
        \protect\label{equation:kinematic_2}
\end{equation}
with matrix ${\bf E}$ defined as
\begin{eqnarray}
{\bf E}= \left[\begin{array}{cc}
              0 & -1 \\
              1 &  0
             \end{array}
        \right]                  \nonumber
\end{eqnarray}
\end{subequations}
and \negr c and \negr d denoting the position vectors, in the frame
indicated in Fig.~\ref{figure:manipulateur_general}, of points $C$
and $D$, respectively.
\par
Furthermore, note that $\dot{\negr c}$ and
$\dot{\negr d}$ are given by
\begin{eqnarray}
\dot{\negr c}= \dot{\theta}_1 {\bf E}\negr c, \quad\dot{\negr d}=
\dot{\theta}_2 {\bf E}(\negr d - \negr b)   \nonumber
\end{eqnarray}
We would like to eliminate the two idle joint rates $\dot{\theta}_3$ and
$\dot{\theta}_4$ from eqs.(\ref{equation:kinematic_1}) and
(\ref{equation:kinematic_2}), which we do upon dot-multiplying the
former by $\negr p-\negr c$ and the latter by $\negr p-\negr d$, thus
obtaining
\begin{subequations}
\begin{equation}
      ({\bf p - c})^T \dot{\bf p} = ({\bf p} - {\bf c})^T \dot{\bf c}
      \protect\label{equation:kinematic_3}
\end{equation}
\begin{equation}
      ({\bf p - d})^T \dot{\bf p} = ({\bf p} - {\bf d})^T \dot{\bf d}
      \protect\label{equation:kinematic_4}
\end{equation}
\end{subequations}
Equations (\ref{equation:kinematic_3}) and
(\ref{equation:kinematic_4}) can now be cast in vector form, namely,
\begin{subequations}
\begin{equation}
{\bf A} \dot{\negr p}={\bf B \dot{\gbf\theta}}\label{e:Adp=Bdth}
\end{equation}
with $\dot{\gbf{\theta}}$ defined as the vector of actuated joint
rates, of components $\dot{\theta}_1$ and $\dot{\theta}_2$.
Moreover ${\bf A}$  and \negr B are, respectively, the
direct-kinematics and the inverse-kinematics matrices of the
manipulator,  defined as
\begin{equation}
{\bf A}= \left[\begin{array}{c}
                ({\bf p}  - {\bf c})^T \\
                ({\bf p}  - {\bf d})^T
              \end{array}
         \right]
       \protect\label{equation:jacobian_matrices_A}
\end{equation}
and
\begin{equation}
{\bf B}= L_1 L_2  \left[\begin{array}{cc}
                          \sin(\theta_3 - \theta_1) &
                          0                         \\
                          0                         &
                          \sin(\theta_4 - \theta_2)
                        \end{array}
                  \right]
       \protect\label{equation:jacobian_matrices_B}
\end{equation}
\end{subequations}
\section{The Isoconditioning Curves}\label{s:iso-curves}
We derive below the loci of equal condition number of the direct- and
inverse-kinematics matrices. To do this, we first recall the
definition of {\em condition number} of an $m \times n$ matrix ${\bf
M}$, with $m \leq n$, $\kappa(\negr M)$. This number can be defined in
various ways; for our purposes, we define $\kappa(\negr M)$ as the
ratio of the largest, $\sigma_l$, to the smallest $\sigma_s$, singular
values of {\bf M}, namely,
\begin{equation}
\kappa({\bf M})= \frac{\sigma_l}{\sigma_s}
\end{equation}
The singular values $\{\sigma_k\}_1^m$ of matrix ${\bf M}$ are defined,
in turn, as the square roots of the nonnegative eigenvalues
of the positive-semidefinite $m \times m$ matrix
${\bf MM^T}$.
\subsection{Direct-Kinematics Matrix}
To calculate the condition number of matrix ${\bf A}$, we need the
product $\negr A\negr A^T$, which we calculate below:
\begin{equation}
{\bf A A}^T = L^2_2 \left[\begin{array}{cc}
                                      1                          &
                                      \cos(\theta_3 - \theta_4)  \\
                                      \cos(\theta_3 - \theta_4)  &
                                      1
                    \end{array}
                    \right]
\end{equation}
The eigenvalues $\alpha_1$ and $\alpha_2$ of the above product are given
by:
\begin{equation}
\alpha_1= 1 - \cos(\theta_3 - \theta_4),\;\;
\alpha_2= 1 + \cos(\theta_3 - \theta_4)
\end{equation}
and hence, the condition number of matrix ${\bf A}$ is
\begin{equation}
\kappa({\bf A})= \sqrt{
                    \frac{\alpha_{max}}{\alpha_{min}}
                      } \\
\end{equation}
where
\begin{equation}
   \alpha_{min}= 1 - \left|\cos(\theta_3 - \theta_4)\right|,
   \quad
   \alpha_{max}= 1 + \left|\cos(\theta_3 - \theta_4)\right|
\end{equation}
\par
Upon simplification,
\begin{equation}
       \kappa({\bf A})= \frac{1}{|\tan((\theta_3 - \theta_4) / 2)|}
       \protect\label{equation:condionning_A}
\end{equation}
In light of expression (\ref{equation:condionning_A}) for the condition
number of the Jacobian matrix {\bf A}, it is apparent that
$\kappa({\bf A})$ attains its  minimum of $1$ when
$\left| \theta_3 - \theta_4 \right|= \pi / 2$,
the equality being understood {\em modulo}
$\pi$. At the other end of the spectrum,
$\kappa({\bf A})$ tends to infinity when $\theta_3 - \theta_4= k
\pi$, for $k=1, 2, \ldots$. When matrix \negr A attains a condition
number of unity, it is termed {\em isotropic}, its inversion being
performed without any roundoff-error amplification. Manipulator
postures for which condition $\theta_3 - \theta_4= \pi/2$ holds are
thus the most accurate for purposes of the direct kinematics of the
manipulator. Correspondingly, the locus of points whereby matrix \negr
A is isotropic is called the {\em isotropy locus} in the Cartesian
workspace.
\par
On the other hand, manipulator postures whereby $\theta_3
- \theta_4= k\pi$ denote a singular matrix \negr A.
Such singularities occur at the boundary of the Joint space of the
manipulator, and hence, the locus of $P$ whereby these
singularities occur, namely, the {\em singularity locus} in the
Joint space, defines this boundary. Interestingly, isotropy can be
obtained regardless of the dimensions of the manipulator, as long
as $i$) it is symmetric and $ii$) $L_2 \neq 0$.
\subsection{Inverse-Kinematics Matrix}
By virtue of the diagonal form of matrix \negr B, its singular values,
$\beta_1$ and $\beta_2$, are simply the absolute values of its
diagonal entries, namely,
\begin{equation}
\beta_1=| \sin(\theta_3 - \theta_1)|,\;\; \beta_2= |\sin(\theta_4 -
\theta_2)|
\end{equation}
The condition number $\kappa$ of matrix ${\bf B}$ is thus
\begin{equation}
\kappa({\bf B})= \sqrt{
                    \frac{\beta_{\rm max}}{\beta_{\rm min}}
                    } \\
       \protect\label{equation:condionning_B}
\end{equation}
where, if $|\sin(\theta_3 - \theta_1)| < |\sin(\theta_4 - \theta_2)|$, then
\begin{equation}
\beta_{\rm min}= |\sin(\theta_3 - \theta_1)|,\;\;
\beta_{\rm max}= |\sin(\theta_4 - \theta_2)|\, ;
\end{equation}
else,
\begin{equation}
\beta_{\rm min}= |\sin(\theta_4 - \theta_2)|,\;\;
\beta_{\rm max}= |\sin(\theta_3 - \theta_1)|\, .
\end{equation}
In light of expression (\ref{equation:condionning_B}) for the condition
number of the Jacobian matrix {\bf B}, it is apparent that
$\kappa({\bf B})$ attains its  minimum of $1$
when $\left|\sin( \theta_3 - \theta_1) \right|= \left|\sin( \theta_4 -
\theta_2)\right|\ne 0$. The locus of points where $\kappa(\negr B)=1$,
and hence, where \negr B is isotropic, is called the {\em isotropy
locus} of the manipulator in the joint space. At the other end of
the spectrum, $\kappa({\bf B})$ tends to infinity when $|\theta_3 -
\theta_1|= k \pi$ or $|\theta_4 - \theta_2|= k \pi$, for $k=1, 2,
\ldots$, which denote singularities of {\bf B}. These singularities
are associated with the inverse kinematics of the manipulator, and
hence, lie within its Cartesian workspace, not at the boundary of
this one. The singularity locus of \negr B thus defines the
Cartesian workspace of the manipulator. Therefore, the Cartesian
workspace of the manipulator is bounded by the singularity locus of
\negr B, i.e., the locus of points where $\kappa({\bf B})
\rightarrow
\infty$. Interestingly, \negr B can be rendered isotropic
regardless of the dimensions of the manipulator, as long as $i$) it
is symmetric and $ii$) $L_1 \neq 0$ and $L_2 \neq 0$.
\subsection{The Working Mode}
\begin{figure}[hbt]
  \begin{center}
        \includegraphics[width= 80mm,height= 60mm]{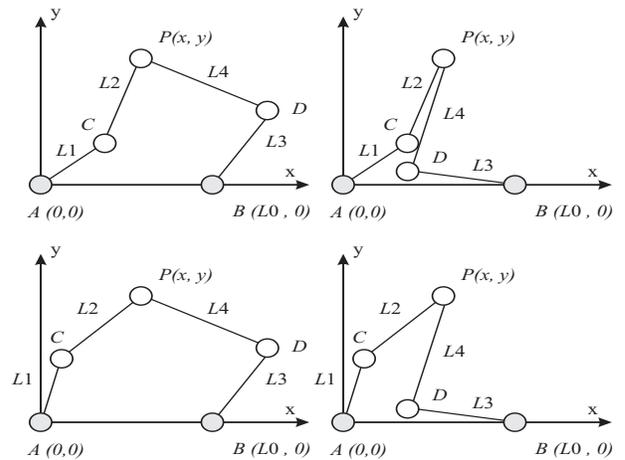}
        \caption{The four working modes}
        \protect\label{figure:working_mode}
  \end{center}
\end{figure}
The manipulator under study has a diagonal inverse-kinematics matrix
\negr B, as shown in eq.(\ref{equation:jacobian_matrices_B}), the
vanishing of one of its diagonal entries thus indicating the
occurrence of a {\em serial singularity}. The set of manipulator
postures free of this kind of singularity is termed a {\em working
mode}. The different working modes are thus separated by a serial
singularity, with a set of postures in different working modes
corresponding to an inverse kinematics solution.
\par
The formal definition of the working mode is detailed in
\cite{Chablat:97}. For the manipulator at hand, there are four
working modes, as depicted in Fig.~\ref{figure:working_mode}.
\subsection{Examples}
We assume here the dimensions $L_0=6$, $L_1=8$, and $L_2=5$, in
certain units of length that we need not specify.
\begin{figure}[hbt]
    \begin{center}
    \begin{tabular}{cc}
       \begin{minipage}[t]{40 mm}
           \centerline{\hbox{\includegraphics[width= 40mm,height= 40mm]{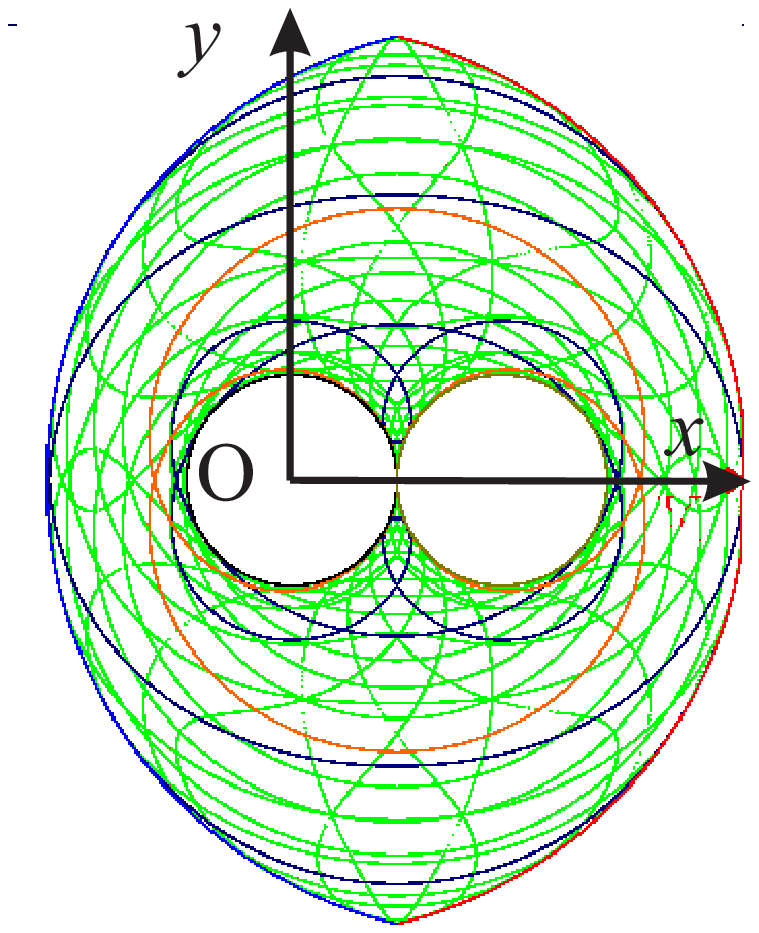}}}
           \caption{The isoconditioning curves in the Cartesian space}
           \protect\label{figure:isoconditionning_curves_workspace}
       \end{minipage} &
       \begin{minipage}[t]{40 mm}
           \centerline{\hbox{\includegraphics[width= 40mm,height= 40mm]{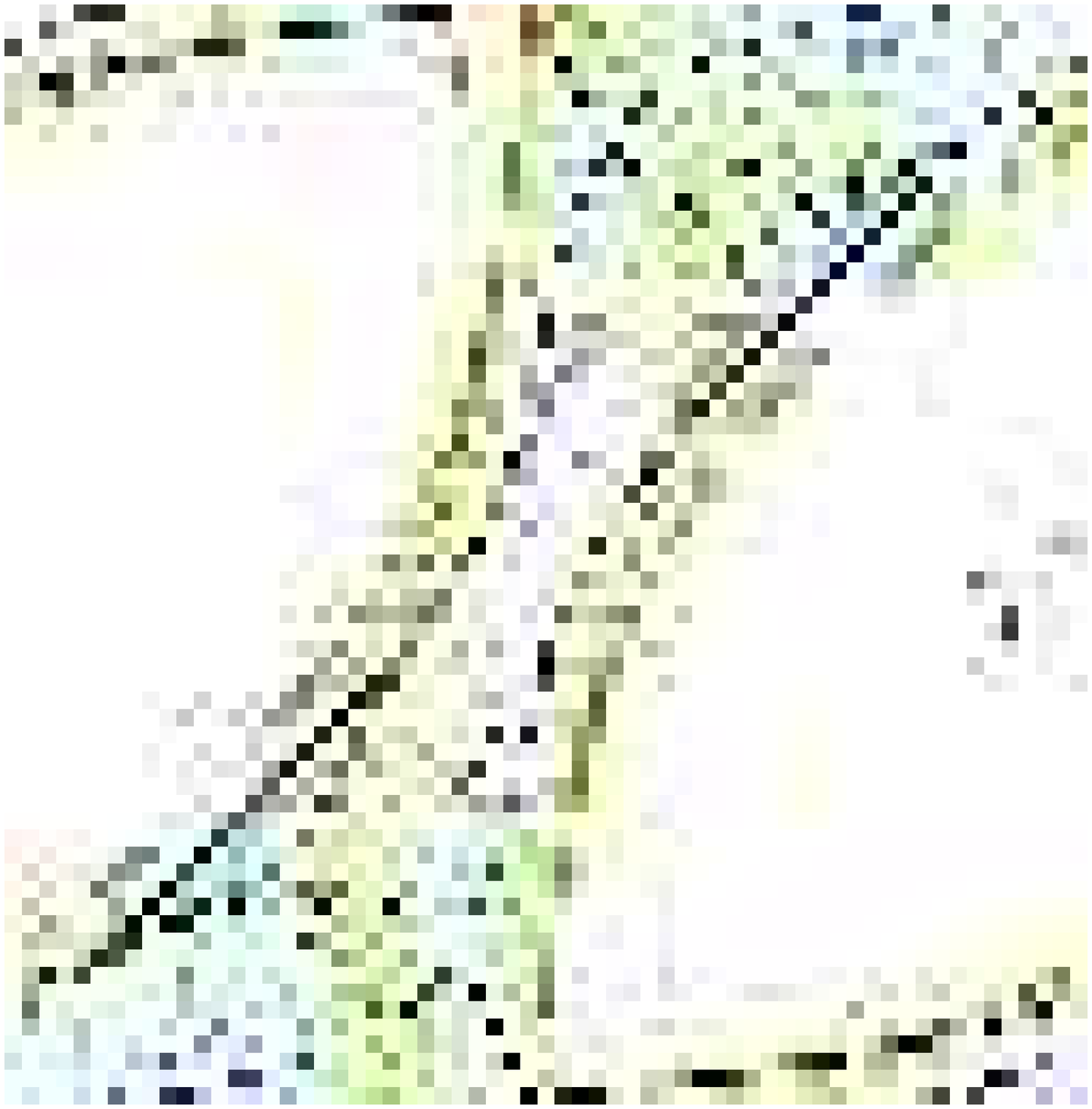}}}
           \caption{The isoconditioning curves in the joint space}
           \protect\label{figure:isoconditionning_curves_jointspace}
       \end{minipage}
    \end{tabular}
    \end{center}
\end{figure}
\par
\begin{figure}
     \begin{center}
     \begin{tabular}{cc}
                  {\includegraphics[width= 35mm,height= 35mm]{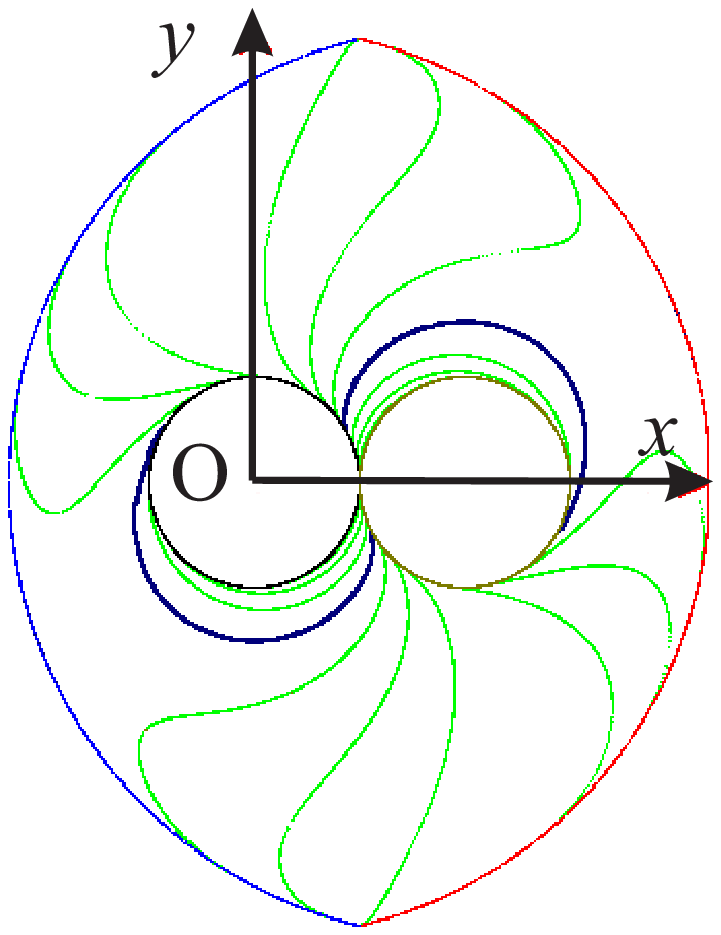}} &
                  {\includegraphics[width= 35mm,height= 35mm]{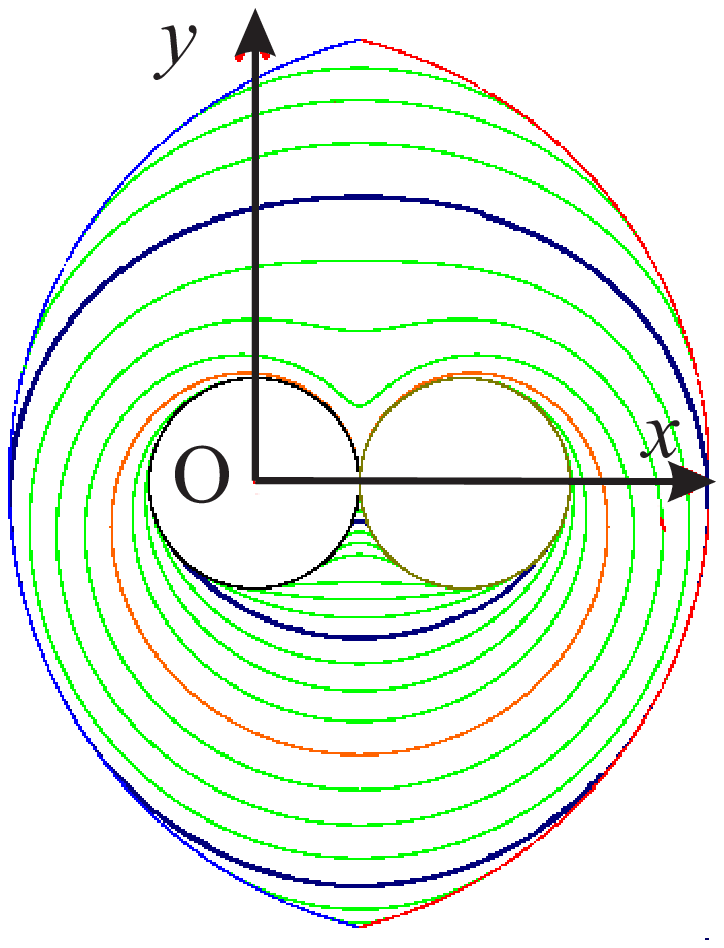}} \\
                  {\includegraphics[width= 35mm,height= 35mm]{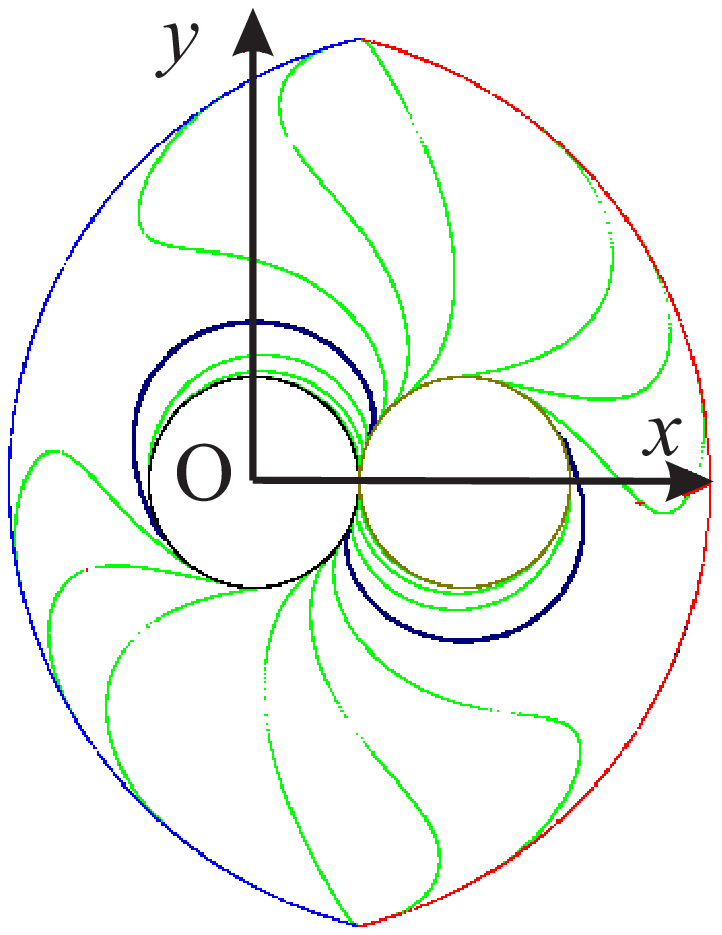}} &
                  {\includegraphics[width= 35mm,height= 35mm]{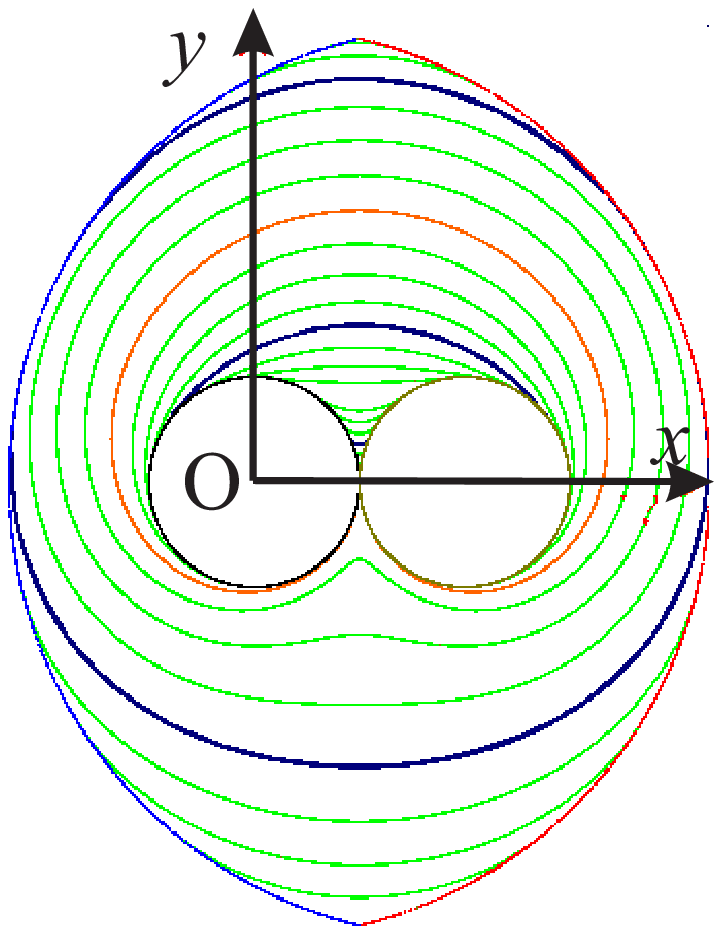}}
     \end{tabular}
     \end{center}
     \caption{The four working modes and their isoconditioning
       curves in the Cartesian space}
     \label{figure:four_working_modes_in_workspace}
\end{figure}

The isoconditioning curves for the direct-kinematic matrix both
in the Cartesian and in the joint spaces are displayed in
Figs.~\ref{figure:isoconditionning_curves_workspace} and
\ref{figure:isoconditionning_curves_jointspace}, respectively.
A better representation of isoconditioning curves can be obtained
in the Cartesian space by displaying these curves for every working
mode, which we do in
Fig.~\ref{figure:four_working_modes_in_workspace}.
\par
In this figure, the isoconditioning curves are the coupler curves of
the four-bar linkage derived upon locking the middle joint, of center
$P(x,\,y)$, to yield a fixed value of $\theta_3-\theta_4$.
Each configuration where points $C$ and $D$ coincide leads to a
singularity where the position of point $P$ is not controllable.
\section{A Three-DOF Hybrid Manipulator}
\begin{figure}[hbt]
  \begin{center}
        \includegraphics[width= 40mm,height= 45mm]{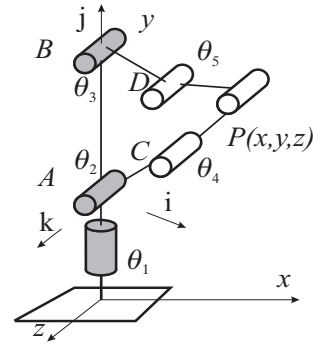}
        \caption{The three-dof hybrid manipulator}
        \protect\label{figure:manipulateur_hybrid}
  \end{center}
\end{figure}
Now we add one-dof to the manipulator of
Fig.~\ref{figure:manipulateur_general}. We do this by allowing the
overall two-dof manipulator to rotate about line $AB$ by means
of a revolute coupling the fixed link of the above manipulator with
the base of the new manipulator. We thus obtain the manipulator of
Fig.~\ref{figure:manipulateur_hybrid}.
\subsection{Kinematic Relations}
The velocity $\dot{\negr p}$ of point $P$ can be obtained in two
different forms, depending on the direction in which the loop is
traversed, namely,
\begin{subequations}
\be
   \dot{\bf p} = \dot{\negr c} + (\dot{\theta}_1 {\bf j} + \dot{\theta}_4
   {\bf k}) \times ({\bf p - c})\label{e:p-c}
\ee
and
\be
  \dot{\bf p} = \dot{d} +
                (\dot{\theta}_1 {\bf j} + \dot{\theta}_5 {\bf k})
                \times ({\bf p - d})
  \label{e:p-d}
\ee
\end{subequations}
\par
Upon dot-multiplying eq.(\ref{e:p-c}) by $(\negr p-\negr c)$ and
eq.(\ref{e:p-d}) by $(\negr p-\negr d)$, we obtain two scalar equations
free of $\dot{\theta}_1$ and the idle joint rates $\dot{\theta}_4$ and
$\dot{\theta}_5$, i.e.,
\beqa
 ({\bf p - c})^T \dot{\bf p}& =& ({\bf p} - {\bf c})^T \dot{\bf c}
        \protect\label{equation:kinematic_3_dof_1}\\
 ({\bf p - d})^T \dot{\bf p}& =& ({\bf p} - {\bf d})^T \dot{\bf d}
        \protect\label{equation:kinematic_3_dof_2}
\eeqa
Furthermore, we note that $\dot{\negr c}$ and $\dot{\negr d}$ are
given by
\beqa
     \dot{\bf c}&=& (\dot{\theta}_1 {\bf j} + \dot{\theta}_2 {\bf k})
        \times {\bf c}
     \label{e:c-dot}\\
     \dot{\bf d}&=& (\dot{\theta}_1 {\bf j} + \dot{\theta}_3 {\bf k} )
        \times({\bf d - b})
     \label{e:d-dot}
\eeqa
Substitution of the above two equations into
eqs.(\ref{equation:kinematic_3_dof_1} \&
\ref{equation:kinematic_3_dof_2}), two kinematic relations between
joint rates and Cartesian velocities are obtained, namely,
\beqa
    \left[({\bf p} - {\bf c}) \times {\bf c}\right]
    \cdot {\bf k}\dot{\theta}_2&=&({\bf p - c})^T \dot{\bf p}
    \label{e:dpkdth_2}\\
    \left[({\bf p} - {\bf d}) \times ({\bf d - b})\right]
    \cdot {\bf k}\dot{\theta}_3&=&({\bf p - d})^T \dot{\bf p}
    \label{e:dpkdth_3}
\eeqa
Moreover, upon dot-multiplying eqs.(\ref{e:p-c} \& b) by \negr k,
we obtain two expressions for the projection of $\dot\negr p$ onto
the $Z$ axis
\beqa
  \negr k^T \dot{\bf p} &=& \negr k^T \left[
  \dot{\negr c} + \dot{\theta}_1 {\bf j} \times ({\bf p - c})\right]
  \nonumber\\
  \negr k^T \dot{\bf p} &=& \negr k^T \left[
  \dot{\negr d} + \dot{\theta}_1 {\bf j} \times ({\bf p - d})\right]
  \nonumber
\eeqa
which, in light of eqs.(\ref{e:c-dot} \& \ref{e:d-dot}), readily reduce to
\beqa
  \negr k^T\dot{\negr p} &=& {\bf i^T p} \dot{\theta}_1 \nonumber\\
  \negr k^T\dot{\negr p} &=& {\bf i^T p} \dot{\theta}_1 \nonumber
\eeqa
It is apparent that the right-hand sides of the two foregoing
equations are identical, and hence, those two scalar equations lead to
exactly the same relation, namely,
\bed
  \negr k^T\dot{\negr p} = ({\bf i^T p}) \dot{\theta}_1 \nonumber
\eed
It will prove useful to have the two sides of the above equation
multiplied by $L_2$, and hence, that equation is equivalent to
\be
  L_2\negr k^T\dot{\negr p} = L_2(\negr i^T \negr p)\dot{\theta}_1
  \label{e:third-eqn}
\ee
In the next step, we assemble eqs.(\ref{e:dpkdth_2} \& \ref{e:dpkdth_3}),
which leads
to an equation formally identical to eq.(\ref{e:Adp=Bdth}), but with
\negr A and \negr B defined now as $3\times 3$ matrices, i.e.,
\begin{subequations}
\beqa
     \!\!\!\negr A\!\!\!\!\!\!&\equiv &\!\!\!\!\!\!
     \left[\begin{array}{c}
           L_2\negr k^T \\
           (\negr p-\negr c)^T \\
           (\negr p-\negr d)^T
           \end{array}
     \right]
     \label{e:A-3dof}\\
     \!\!\!\negr B\!\!\!\!\!\!&\equiv& \!\!\!\!\!\!L_1L_2
     \left[\begin{array}{ccc}
           \!\!\!\!\sin\theta_2+\lambda_1\sin\theta_4\!\!\!\!\!&
           0                                                   &
           0                                                   \\
           0                                                   &
           \!\!\!\!\!\sin(\theta_2-\theta_4) \!\!\!\!\!\!\!    &
           0                                                   \\
           0                                                   &
           0                                                   &
           \!\!\!\!\!\sin(\theta_3-\theta_5) \!\!
           \end{array}
     \right]
     \label{e:B-3dof}
\eeqa
\end{subequations}
with $\lambda_1$ defined as $\lambda_1\equiv L_2/L_1$,
while vectors $\dot{\gbf\theta}$ and $\dot\negr p$ are now
given by
\be
\dot{\gbf\theta}\equiv  \left[\begin{array}{c}
                                 \dot{\theta}_1 \\
                               \dot{\theta}_2 \\
                                 \dot{\theta}_3
                              \end{array}
                        \right],\quad
\dot{\negr p}\equiv\    \left[\begin{array}{c}
                               \dot{x} \\
                               \dot{y} \\
                               \dot{z}
                              \end{array}
                        \right]\quad
\label{e:dtheta&dp}
\ee
\section{The Isoconditioning Surfaces}
We conduct here the same analysis of Section~\ref{s:iso-curves}.
\subsection{The Direct-Kinematics Matrix}
\begin{figure}[hbt]
        \begin{center}
        \includegraphics[width= 50mm,height= 50mm]{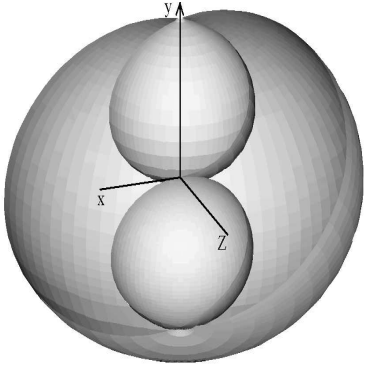}
        \caption{The boundary of the workspace}
        \protect\label{figure:workspace_3d}
        \end{center}
\end{figure}

Apparently, matrix \negr A in the 3-dof case has a structure similar
to the corresponding matrix in the 2-dof case. Indeed, upon
calculating $\negr A\negr A^T$ in the 3-dof case, we obtain
\be
    \negr A\negr A^T= L_2^2
        \left[\begin{array}{ccc}
     1  &  0                      &  0                       \\
     0  &  1                      &  \cos(\theta_4-\theta_5) \\
     0  & \cos(\theta_4-\theta_5) & 1
              \end{array}
        \right]
    \label{e:AA^T-3dof}
\ee
The eigenvalues of the foregoing matrix are, then,
$\alpha_1=1-|\cos(\theta_4-\theta_5)|$, $\alpha_2=1$, and
$\alpha_3=1+|\cos(\theta_4-\theta_5)|$, the foregoing eigenvalues
having been ordered as
\bed
        \alpha_1\le \alpha_2\le\alpha_3
\eed
The condition number of matrix \negr A is thus
\bed
  \kappa(\negr A)=\sqrt{\frac{1+|\cos(\theta_4-\theta_5)|}
  {1-|\cos(\theta_4-\theta_5)|}}
\eed
which can be further simplified to
\be
    \kappa(\negr A)=\frac{1}{|\tan((\theta_4-\theta_5) / 2)|}
    \label{e:kappa(A)-3dof}
\ee
Therefore, the condition number of the two direct-kinematics
matrices, for the 2-dof and the 3-dof cases, coincide. However, the
loci of isoconditioning points are now surfaces, because we have
added one dof to the manipulator of
Fig.~\ref{figure:manipulateur_general}. These loci are, in fact,
surfaces of revolution generated by the isoconditioning curves of
the 2-dof manipulator, when these are rotated about the axis of the
first revolute. We represent the boundary of the workspace (Fig.
\ref{figure:workspace_3d}).
\subsection{The Inverse-Kinematics Matrix}
Given the diagonal structure of matrix \negr B, its singular values
are apparently, $\{\,L_1L_2\beta_i\,\}_1^3$, with the definitions
below:
\beqa
    \beta_1&=&|\sin\theta_2+\lambda_1\sin\theta_4|,\nonumber\\
    \beta_2&=&|\sin(\theta_2-\theta_4)|, \nonumber \\
    \beta_3&=&|\sin(\theta_3-\theta_5)|\nonumber
\eeqa
Therefore, the isoconditioning locus of \negr B is determined by the
relation
\be
    |\sin\theta_2+ \lambda_1\sin\theta_4|=
    |\sin(\theta_2-\theta_4)|=
    |\sin(\theta_3-\theta_5)|\nonumber
    \label{e:Bisocond}
\ee
Notice that the distance $d_1$ of $P$ to the $Y$ axis is
\be
   d_1=L_1 \sin\theta_2 + L_2 \sin\theta_4=L_1\beta_1
   \label{e:d_1}
\ee
Likewise, the distances $d_2$ and $d_3$ of $P$ to the two axes of the
other two actuated revolutes, i.e., those passing through $A$ and $B$
are, respectively,
\beqa
d_2&=&L_2\beta_2\label{e:d_2}\\
d_3&=&L_2\beta_3\label{e:d_3}
\eeqa
It is now straightforward to realize that, for the case at hand, the
locus of isotropic points of \negr B are given by manipulator postures
whereby $P$ is equidistant from the three actuated revolute
axes. Likewise, postures whereby point $P$ lies on the $Y$ axis are
singular; at these postures, $\kappa (\negr B)$ tends to infinity.
Moreover, the inverse-kinematics singularities occur whenever any of
the diagonal entries of \negr B vanishes, i.e., when
\beqa
   d_1=0,\;{\rm or } \; \theta_2=\theta_4+k\pi,\;{\rm or } \;
   \theta_3=\theta_5+k\pi\label{e:Bsingular}
\eeqa
for $k=1,2,\ldots$.
\section{Conclusions}
We have defined a new architecture of hybrid manipulators and derived
the associated loci of isoconditioning points. Two Jacobian matrices
were identified in the mapping of joint rates into Cartesian
velocities, namely, the direct-kinematics and the inverse-kinematics
matrices. Isoconditioning loci were defined for these matrices. Two
special loci were discussed, namely, those pertaining to isotropy and
to singularity, for each of these matrices.
\par
The study has been conducted for three-dof-hybrid manipulators but
applies to six-dof-hybrid manipulators with wrist as well.
\par
The hybrid manipulators studied have interesting features like
workspace and high dynamic performances, which are usually met
separately in serial or parallel manipulators, respectively. Futher
research work is being conducted by the authors on such hybrid
manipulators with regard to their optimal design.
\section*{Acknowledgments}
The third author acknowledges the support from the Natural Sciences
and Engineering Research Council, of Canada, the Fonds pour la
formation de chercheurs et l'aide \`a la recherche, of Quebec, and
\'Ecole Centrale de Nantes (ECN). The research reported here was conducted
during a sojourn that this author spent at ECN's Institut de Recherche
en Cybern\'etique de Nantes.
\bibliographystyle{unsrt}

\end{document}

%% file: macro.tex
\newcommand{\gbf}[1] {\mbox{\boldmath${#1}$\unboldmath}}
\newcommand{\be}{\begin{equation}}
\newcommand{\ee}{\end{equation}}
\newcommand{\beq}{\begin{equation}}
\newcommand{\eeq}{\end{equation}}
\newcommand{\bed}{\begin{displaymath}}
\newcommand{\eed}{\end{displaymath}}
\newcommand{\beqa}{\begin{eqnarray}}
\newcommand{\eeqa}{\end{eqnarray}}
\newcommand{\beqann}{\begin{eqnarray*}}
\newcommand{\eeqann}{\end{eqnarray*}}
\newcommand{\bseq}{\begin{subequation}}
\newcommand{\eseq}{\end{subequation}}

\newcommand{\ba}{\begin{array}}
\newcommand{\ea}{\end{array}}

\newcommand{\1}{{\bf 1}}

\newcommand{\negr}[1]{{\bf {#1}}}

%% file: icra_Chablat_Wenger_bis.bbl
\begin{thebibliography}{99}
\bibitem{Hunt:78}
Hunt, K.~H.
\newblock ``Geometry of Mechanisms''
\newblock Clarendon Press, Oxford, 1978.
\bibitem{Bajpai:86}
Bajpai, A.\ and Roth, B.
\newblock ``Workspace and mobility of a closed-loop manipulator''
\newblock The International Journal of Robotics Research, Vol.~5, No.~2,
1986.
\bibitem{Gosselin:90b}
Gosselin, C. 
\newblock ``Stiffness mapping for parallel manipulators''
\newblock IEEE Transactions On Robotics And Automation, Vol.~6, No.~3,
June 1990.
\bibitem{Gosselin:90a}
Gosselin, C.\ and Angeles, J.
\newblock ``Singularity analysis of closed-loop kinematic chains''
\newblock IEEE Transactions On Robotics And Automation, Vol.~6, No.~3,
June 1990.
\bibitem{Chablat:97}
Chablat, D. and Wenger, Ph.
\newblock``Working modes and aspects in fully parallel manipulators''
\newblock to appear in Proc. IEEE International Conference of Robotic and Automation,
Mai 1998.
\bibitem{Golub:89}
Golub, G.~H.\ and Van Loan, C.~F.
\newblock ``Matrix Computations''
\newblock The Johns Hopkins University Press, Baltimore, 1989.
\end{thebibliography}
